\documentclass{statsoc}

\newcommand{\balpha}{ \mbox{\boldmath $\alpha$}}
\newcommand{\bbeta}{ \mbox{\boldmath $\beta$}}

\newcommand{\bx}{ \mbox{\bf x}}

\newcommand{\bs}{ \mbox{\bf s}}

\newcommand{\bv}{ \mbox{\bf v}}

\newcommand{\iid}{\stackrel{iid}{\sim}}

\newcommand{\calD}{{\cal D}}

\newcommand{\calP}{{\cal P}}

\renewcommand{\and}{\quad\text{and}\quad}

\newcommand{\ind}[3][1]{#2 = #1,\ldots,#3}

\newcommand{\where}{\quad\text{where}\quad}

\newcommand{\given}{\,|\,}

\newcommand{\beq}{ \begin{equation}}
\newcommand{\eeq}{ \end{equation}}
\newcommand{\beqn}{ \begin{eqnarray}}
\newcommand{\eeqn}{ \end{eqnarray}}

\usepackage[a4paper]{geometry}
\usepackage{graphicx}
\usepackage{amsmath}
\usepackage{natbib}
\usepackage{hyperref}

\title[Global forensic geolocation with deep neural networks]{Global forensic geolocation with deep neural networks}
\author{Neal S. Grantham, Brian J. Reich, Eric B. Laber,}
\address{Department of Statistics, North Carolina State University}%, Raleigh, NC 27695}
\author{Krishna Pacifici,}
\address{Department of Forestry and Environmental Resources, North Carolina State University}%, Raleigh, NC 27695}
\author{Robert R. Dunn,}
\address{Department of Applied Ecology, North Carolina State University}%, Raleigh, NC 27695}

\author{Noah Fierer, Matthew Gebert,}
\address{Department of Ecology and Evolutionary Biology, University of Colorado}%, Boulder, CO 80309}
\author[Grantham et al.]{Julia S. Allwood and
Seth A. Faith}
\address{Forensic Science Institute, North Carolina State University}%, Raleigh, NC 27695}

\begin{document}
\begin{abstract}
An important problem in forensic analyses is identifying the provenance of materials at a crime scene, such as biological material on a piece of clothing. This procedure, known as geolocation, is conventionally guided by expert knowledge of the biological evidence and therefore tends to be application-specific, labor-intensive, and subjective.  Purely data-driven methods have yet to be fully realized due in part to the lack of a sufficiently rich data source. However, high-throughput sequencing technologies are able to identify tens of thousands of microbial taxa using DNA recovered from a single swab collected from nearly any object or surface. We present a new algorithm for geolocation that aggregates over an ensemble of deep neural network classifiers trained on randomly-generated Voronoi partitions of a spatial domain. We apply the algorithm to fungi present in each of 1300 dust samples collected across the continental United States and then to a global dataset of dust samples from 28 countries. Our algorithm makes remarkably good point predictions with more than half of the geolocation errors under 100 kilometers for the continental analysis and nearly 90\% classification accuracy of a sample's country of origin for the global analysis. We suggest that the effectiveness of this model sets the stage for a new, quantitative approach to forensic geolocation.
\end{abstract}

\section{Introduction}

Forensic geolocation is the act of identifying the geographic source of an item based on its observable properties. For example, in the course of a criminal investigation, soil on a suspect might be compared to soil at a crime scene \citep[e.g.,][]{Pye}. Another promising source of evidence for use in geolocation is the dust that accumulates on buildings, objects, and individuals.  The analysis of dust traces to aid forensic inquiry is not a new idea. In the 1920s, during his tenure as Director of the Laboratory of Police Technique in Lyon, France, Edmond Locard expressed excitement about the potential impact of dust-based forensic work, remarking ``the microscopic debris that cover our clothes and bodies are the mute witnesses, sure and faithful, of all our movements and of all our encounters'' \citep{locard1930analysis}.
	
Realization of Locard's vision, however, has been slow and episodic, particularly with regard to the biological materials in dust. A primary reason is that dust analysis has historically relied on the identification of microscopic pollen grains or other biological materials (a laborious task) \citep{bryant2006forensic}. Such work is clearly useful, but inherently subjective and highly dependent on the skill and perspective of the experts.  Recent work has expanded the capabilities of pollen for forensic analysis \citep{jones2012forensic,goodman2015piglt}, but the technique remains used more often in archeology than for forensic geolocation.
	
Dust can harbor DNA from bacteria, archaea, fungi, plants, and animals and we can leverage recent developments in high-throughput DNA sequencing to assess the composition of these biotic assemblages \citep{barberan2015continental,madden2016diversity,craine2017molecular}. Such DNA analyses can generate a more comprehensive biogeographical fingerprint than is possible with traditional, and widely used, approaches that rely on morphology-based assessments of pollen alone.  Advances in sequencing technology allow for high-fidelity identification of hundreds of thousands of taxa within samples containing biological material.  

Soil forensics is a developing field with a focus on the organic characteristics of soil, rather than the traditional inorganic, physical components,  to reveal clues as to trace material origin or to link an evidentiary item or suspect to a crime scene.  Various DNA-based methods have been investigated to identify the microbial communities within the soil and evaluate their use as spatial signatures and reference points for forensic use \citep{damaso2018bioinformatics,young2015predicting}.  Grantham et al. \cite{grantham2015fungi} developed a preliminary forensic geolocation model for fungal occupancy (presence/absence) data built on Bayesian discriminant analysis (BDA) that compares a sample to a reference database and predicts the sample's provenance.  When applied to data derived from dust samples from homes across the continental United States, the geolocation algorithm makes point predictions that fall within 230 kilometers, on average, of their true provenance. These were the first results to formally estimate the extent to which dust-associated fungi can predict the provenance of a sample.

We use deep learning to obtain more accurate geolocation models from microbiome data.
Deep learning has recently been applied successfully to related high-dimensional problems.  For example, deep neural networks have greatly improved human speech recognition software \citep{hinton2012deep} and they have mastered the ancient board game of Go \citep{silver2016mastering}. In the context of geolocation, \cite{weyand2016planet} have developed the so-called PlaNet that is capable of geolocating photos taken across the globe with accuracy superior to the average human. PlaNet partitions the spatial domain into discrete cells and fits a convolutional neural network to millions of images taken from these cells. A similar approach could prove fruitful for capturing the complexity inherent to high-dimensional microbiome data, but the approach has to be modified to produce point-level geolocation predictions.
	
We demonstrate the utility of DeepSpace, a novel forensic geolocation algorithm which merges features of spatial statistical methods and deep learning to model spatially-distributed point-level data for forensic geolocation using 
microbiome data. In particular, DeepSpace estimates the intensity surface of a spatial point pattern using an ensemble of deep neural network classifiers trained on random Voronoi partitions of the spatial domain. By fitting a flexible deep neural network classifier to each partition and averaging over partitions, the DeepSpace model is sufficiently flexible to approximate any continuous conditional distribution of a sample's provenance given its microbiome composition. We use this new method to analyze the dust fungal microbiome data at the national (continental United States), regional (North Carolina's tri-county Research Triangle), and global (28 countries across 6 continents) scales.  An objective of this analysis is to determine the scales at which geolocation using microbiome data is feasible.  We compare DeepSpace with other geolocation algorithms and find a substantial reduction in cross-validation error at the national level, and, for the first time, demonstrate that microbiome data can be used to successfully geolocate dust samples at the global level.  
	
\section{Microbiome data}\label{sec:data}
	
{\bf Continental U.S. data}: A complete description of the data collection and processing procedures can be found in \cite{barberan2015ecology}, \cite{barberan2015continental}, and \cite{grantham2015fungi} and so we describe these procedures only briefly.  The data were collected by citizen-science volunteers as part of the Wild Life of Our Homes (WLOH) project (homes.yourwildlife.org).  We analyzed dust samples taken by the volunteers from the upper door trim of the outside surface of an exterior door.   Swabs were prepared for sequencing using the PCR approach of \cite{flores2012direct} and sequencing focused on the first internal transcribed spacer (ITS1) region of the rRNA operon using the ITS1-F (CTTGGTCATTTAGAGGAAGTAA) and ITS2 (GCTGCGTTCTTCATCGATGC) primer pair.  Sequences were clustered into phylotypes with greater than or equal to $97\%$ similar sequences.  This dataset includes $n=1,301$ dust samples harboring $p=57,331$ distinct fungal phylotypes.  %The spatial locations of the samples are plotted in Figure \ref{fig:partitions}.  
We also performed a separate analysis of the subset of the samples from Wake, Durham, and Orange Counties in central North Carolina ($n=116$). 
		
{\bf Global data}: In addition to the WLOH analysis, we conduct the first analysis of dust-associated microorganisms collected from 28 countries across 6 continents (excluding Antarctica), the largest database of global dust samples to date. Outdoor dust samples were collected with the Bode SecurSwab 2 collector or a sterile cotton swab between November 2016 and February 2017 by individuals located around the world. The Bode SecurSwab 2 design allowed for the collected samples to be transported back to the U.S. without risk of the swab head touching the sides of the packaging which could have caused sample loss and altered the results.  Among $n=399$ samples, $10-15$ samples were collected from each of the following countries representing several geographic regions: the Americas (Mexico, Costa Rica, Colombia, Trinidad and Tobago, Uruguay, Argentina), Africa (Ghana, Nigeria, South Africa), East Europe (Czechia, Croatia, Hungary, Macedonia), West Asia (Turkey, Cyprus, Jordan), Middle East (Kuwait, Qatar, Oman, Georgia, Azerbaijan), Central Asia (Kazakhstan, Pakistan), East Asia (Vietnam, South Korea, Malaysia), and Oceania (Australia, New Zealand).
	
%\footnotetext{The former Yugoslav Republic of Macedonia (FYROM).}
%\footnotetext{Sample collection in Turkey was split evenly between the cities of Istanbul and Antalya.}
%\footnotetext{Specifically, samples were collected from the northern part of the island of Cyprus, which, while recognized as belonging to the Republic of Cyprus, is de facto controlled by the self-declared Turkish Republic of Northern Cyprus.}
	
	Dust samples were stored desiccated at room temperature and sent to the University of Colorado laboratory for molecular analysis. DNA was extracted using MoBio PowerSoil htp-96 well Isolation Kit and a modified method of \cite{barberan2015continental}.
	To target fungal strains, we amplified the first internal transcribed space (ITS1) region of the rRNA operon using ITS1-F/ITS2 barcoded primers \citep{mcguire2013digging}.
	Each sample was assigned a unique 12-bp error-correcting barcode \citep{caporaso2010qiime} included in the ITS1 primer pairs.
	After PCR amplification in triplicate, the samples were sequenced on an Illumina MiSeq instrument running the 2x250 bp MiSeq kit.
	Sequences were demultiplexed using a custom Python script, pair-end reads were merged using the \texttt{usearch7 mergepairs} feature \citep{edgar2010search}, adapter sequences were trimmed from the merged reads using \texttt{fastx\_clipper} (\url{https://github.com/agordon/fastx_toolkit}), and reads were quality filtered using \texttt{usearch7} \citep{edgar2010search}.
	Finally, sequences were clustered into operational taxonomic units (phylotypes) using the UPARSE pipeline \citep{edgar2013UPARSE} and their taxonomic identities were determined using a Bayesian classifier \citep{wang2007naive} and compared against those in the UNITE database \citep{abarenkov2010UNITE}. 	In total, the extraction, sequencing, and processing pipeline yielded $p=15,475$ distinct fungal phylotypes across $n=399$ samples.

\section{DeepSpace algorithm}\label{sec:algo}

For $j=1,...,J$ we randomly partition the spatial domain $\calD$ and fit a supervised classification algorithm to predict the partition of origin of a sample based on its microbiome; we aggregate these predictions to form a spatial point prediction of the sample's origin.  For random partition $j$, we draw
$K_j$ ``seeds'' from the domain, $\bv_{jk} \iid \text{Uniform}(\calD)$, and define a Voronoi partition $\calP_j=\{P_{jk}\}_{k=1}^{K_j}$ where ``tile'' $j$ is defined as $$P_{jk} = \{\bs\in\calD:||\bs-\bv_{jk}||<||\bs-\bv_{jl}||\;\text{for}\;l\neq k\}.$$ The observation locations are then allocated to the appropriate tile so that $h_j(\bs)=k$ indicates that $\bs\in\calP_{jk}$. For training sample $i=1,...,n$, let $\bs_i$ denote its spatial location and $\bx_i$ the collection of its $p$ OTU presence/absence indicators.  Separately for each $j=1,...,J$, we train a supervised classifier on $\left\{[\bx_i, h_j(\bs_i)]\right\}_{i=1}^n$ to obtain estimates of the probabilities $\mbox{Prob}(\bs\in\calP_{jk}|\bx)=q_{jk}$, denoted ${\hat q}_{jk}(\bx)$. 

Assuming a uniform distribution within each tile and averaging over random partitions gives the predictive distribution
\begin{equation}\label{eq:geolocation}
{\hat g}(\bs\given\bx) = \sum_{j=1}^J\sum_{k=1}^{K_j}\frac{I(\bs\in\calP_{jk}){\hat q}_{jk}(\bx)}{A_{jk}}
\end{equation}
where $A_{jk}$ is the area of $\calP_{jk}$. The spatial prediction for a sample with microbiome data $\bx$ is the location that maximizes the fitted intensity surface,  
$$\hat{\bs}(\bx) = \underset{\bs\in\mathcal{D}}{\arg\max}\;{\hat g}(\bs\given\bx)$$
and $1-\alpha$ prediction regions are computed as $R_{\alpha} = \{\bs: {\hat g}(\bs\given\bx) > T_{\alpha}\}$ for the threshold $T_\alpha$ that gives $\int I[{\hat g}(\bs\given\bx) > T_\alpha]{\hat g}(\bs\given\bx)d\bs \approx \alpha$.

The accuracy of these predictions depends heavily on the performance of the trained classifier that estimates $q_{jk}(\bx)$. There are many possibilities, including classical machine learning classifiers \citep{friedman2001elements} such as $K$-nearest neighbors, support vector machines, random forests; deep learning classifiers \citep{goodfellow2016deep} such as deep neural networks, convolutional neural networks (for image data), recurrent neural networks (for sequential data); or an ensemble of classifiers, where a classifier is selected at random to fit to each new Voronoi partition. Thus, the algorithm accommodates a vast number of possible spatial prediction applications, and the user should choose from classifiers that are appropriate for their data-generating model.
	
Deep learning is well suited for the high-dimensionality and complex dependence structure of our motivating microbial presence/absence data. When paired with a deep learning classifier, such as the deep neural network described below, we call our proposed algorithm DeepSpace for its ability to learn deep patterns in spatially-distributed data. DeepSpace spans a rich class of predictive density functions  $g(\bs\given\bx)$ if we select a nonparametric classifier for the relationship between the covariates and the aggregate probabilities of the $K$ regions defined by a single partition.  Although each of the $J$ estimated functions are discontinuous in space, the model-averaged geolocation function in \eqref{eq:geolocation} can approximate a continuous intensity function as $J$ and $K_j$ increase. Furthermore, if we select a classifier which consistently estimates the aggregate probabilities for the $K$ regions defined by a single partition, then the model-averaged estimator in \eqref{eq:geolocation} should be a consistent nonparametric estimator for the true geolocation function $g(\bs\given\bx)$.
	
A single-layer neural network \citep{goodfellow2016deep} is comprised of $p$ input nodes (input layer), $K$ output nodes (output layer), one for each available class, and $R$ ``hidden'' nodes that connect the nodes of the input layer with those of the output layer. The nodes in the hidden layer are called neurons, as their non-linear transformations of the data are designed to mimic the firing, or ``activation,'' of neurons in the brain when processing sensory data. The single layer neural network 
	\begin{equation}\label{eq:neural_net}
	f_k(\bx) = \beta_{k0} + \sum_{r=1}^R\beta_{kr}\sigma\left(z_r\right)
	\quad\text{where}\quad z_r = \alpha_{r0} + \sum_{j=1}^p\alpha_{rj} x_j,
	\end{equation}
$\sigma(\cdot)$ is a non-linear function, typically chosen to be the inverse logistic function $\sigma(z) = [1 + \exp(-z)]^{-1}$ or the rectified linear unit (ReLU) function $\sigma(z) = \max\{0, z\}$; $\balpha_r = (\alpha_{r0}, \ldots, \alpha_{rp})'$ are unknown coefficients from the regression of neuron $z_r$ on $\bx$ with intercept $\ind{r}{R}$; and $\bbeta_k = (\beta_{k0}, \ldots, \alpha_{kR})'$ are unknown coefficients from regressing $f_k(\bx)$ on activations $\sigma(z_1), \ldots, \sigma(z_R)$ with intercept, $\ind{k}{K}$.
	
Until recently, attempts to introduce more than one hidden layer into a neural network were fraught with convergence difficulties. With the advent of ``pre-training'' methods designed to alleviate these barriers \citep{hinton2006fast}, however, neural networks experienced a renaissance in the form of ``deep learning,'' so named for their newfound ability to learn much richer associations in the data than is possible with a single hidden layer \citep{lecun2015deep}. The single layer neural network \eqref{eq:neural_net} is generalized to a deep neural network by introducing hidden layers $l=1,\ldots,L$ each with $R_l$ neurons so that
$$
f_k(\bx) = \beta_{k0} + \sum_{r=1}^{R_L}\beta_{kr}\sigma\left(z_r^L\right) \where z^l_r = \begin{cases}
\alpha^l_{r0} + \sum\limits_{s=1}^{R_{l-1}}\alpha^l_{rs} \sigma\left(z^{l-1}_s\right) & \ind[2]{l}{L} \\
\alpha_{r0}^1 + \sum\limits_{j=1}^p\alpha^1_{rj} x_j & l=1\\
\end{cases}.$$
The probabilities for each region are obtained from the final output layer, $q_k(\bx) = \exp[f_k(\bs)]/\{\sum_{l=1}^K\exp[f_l(\bs)]\}$. 

We select the categorical cross-entropy function to measure the error of our network, which penalizes the network for assigning high probabilities to incorrect regions. Finally, the backpropagation algorithm and a stochastic gradient-based optimization method, such as Adam \citep{kingma2014adam}, are used to obtain estimates $\hat{q}_1,\ldots,\hat{q}_K$ which minimize the cross-entropy cost over several (10 for the global analysis and 20 for the US analysis) training epochs. 
	
\begin{figure}
\centering
\includegraphics[width=0.85\textwidth]{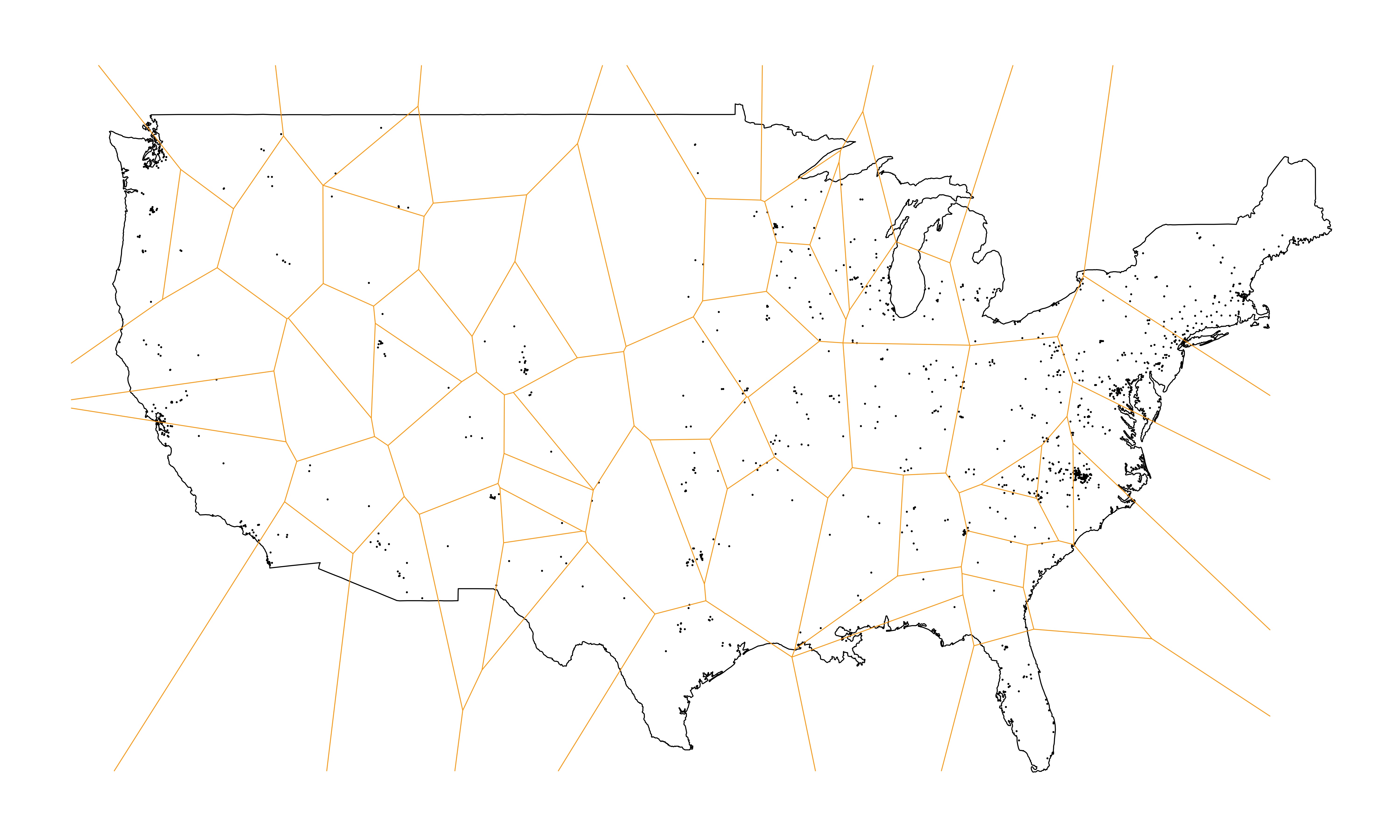}
\includegraphics[width=0.85\textwidth]{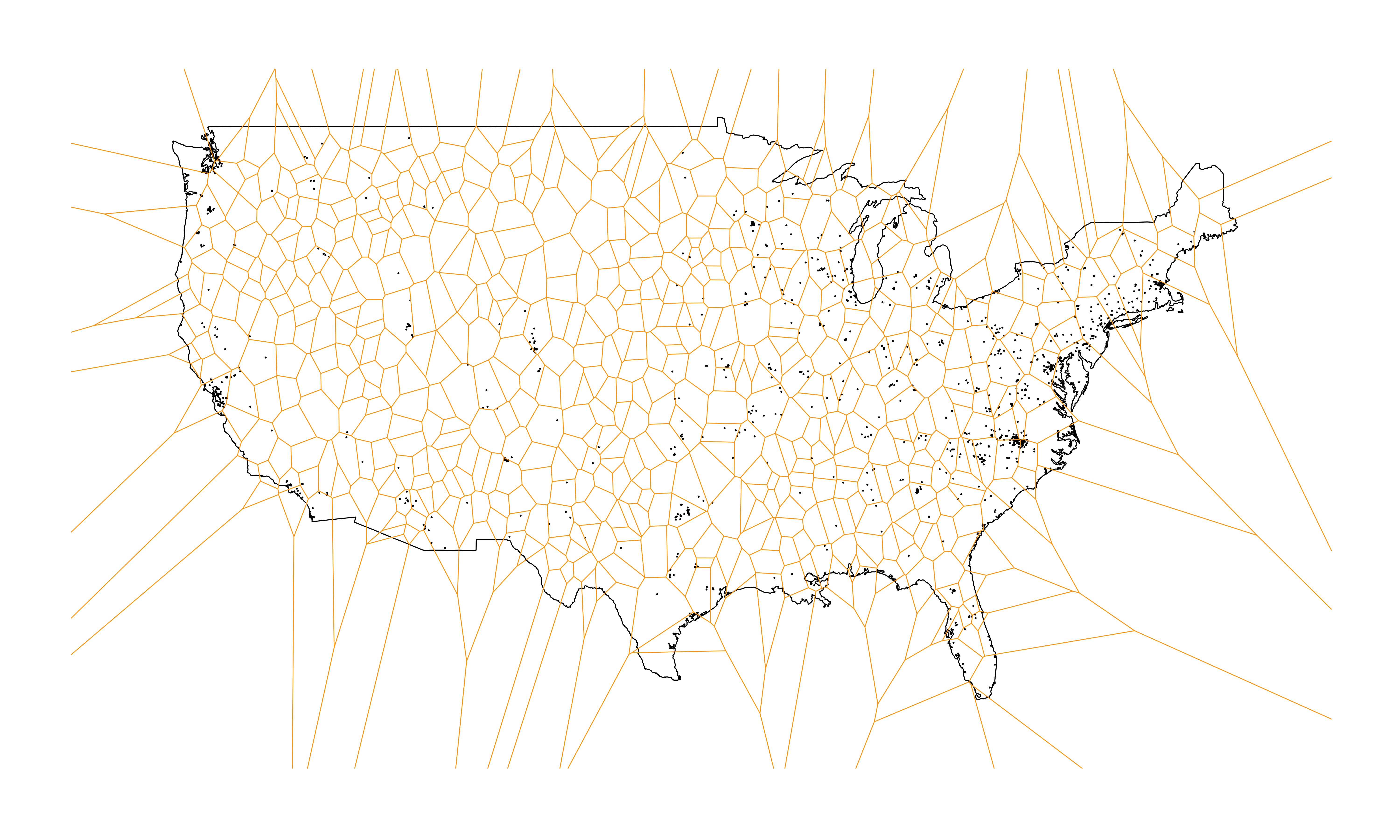}
\caption{Course (top; $K$ equal to 5\% of the sample size) and fine (bottom; $K$ equal to 50\% of the sample size) Voronoi partitions (yellow lines) generated over the continental United States. Sample points (dots) are labeled according to the cell they lie within and a supervised classifier is trained on these labeled data.}
\label{fig:partitions}
\end{figure}

In addition to the tuning parameters in a supervised classification algorithm, DeepSpace requires selecting the number of partition seeds ($K$) and the number of random partitions ($J$).  The number of seeds, $K$, affects the granularity of the Voronoi partitions.  If $K$ is small, then the partitioning is coarse (Figure \ref{fig:partitions}) and most cells are suitably populated, but they may not discriminate well between different regions in $\calD$ and the classifier will infer little about the true spatial variation of the data. If $K$ is large, the partitioning is fine (Figure \ref{fig:partitions}) and most cells lack a suitable number of representative samples for the classifier to learn useful representations of the data from these regions, though it may detect fine-scale variation in well-populated regions. Of course, there is no need to fix the number of seeds for every partition, and one may vary the seed number at the creation of each partition. 

DeepSpace depends on a large number of partition-classifier pairs, $J$, to perform effectively. Although there is no harm in using too many pairs, there is a point of diminishing returns where additional pairs no longer improve predictions.  We use $J=50$ pairs but the results are not sensitive to this choice.

\section{Results}

We compared the following models:
\begin{enumerate}
	\item \textbf{Spatial NN}: Geolocation algorithm using nearest neighbors classifiers with 25 neighbors and the Jaccard distance metric suitable for binary data.
	\item \textbf{Spatial RF}: Geolocation algorithm using random forests classifiers with 200 estimators.
    \item \textbf{Spatial Net}: Geolocation algorithm using neural network classifiers with one hidden layer of 2,048 neurons each with rectified linear unit activation function.
	\item \textbf{DeepSpace (DS)}: Geolocation algorithm using deep neural network classifiers with three hidden layers of 2,048, 1,024, and 1,024 neurons, respectively, each with the ReLU activation function. A dropout rate of 0.3 is applied between the layers of the network to prevent overfitting.
	\item \textbf{Bayesian Discriminant Analysis (BDA)}: The geolocation model developed in \cite{grantham2015fungi}.
	\item \textbf{Area DNN}: A deep neural network classifier built as in DeepSpace, but not placed within the geolocation algorithm; rather, this DNN predicts a sample point's area of origin, i.e., state (national), county (regional), and country (global), and uses the predicted area's centroid as the `most likely' origin.
\end{enumerate}
We further compare several partitioning schemes, including:
\begin{enumerate}
	\item \textbf{Coarse}: $K$ is fixed at approximately 5\% of the sample size, $n$,
	\item \textbf{Fine}: $K$ is fixed at approximately 50\% of $n$.
	\item \textbf{Mixed}: $K$ is allowed to vary and is chosen uniformly to be 5\%, 10\%, ..., 45\%, 50\% of $n$ for each new partition.
    \item \textbf{None}: No seeds are selected (used for the BDA and Area DNN models).
\end{enumerate}
	
Performance at the national, regional, and global levels was measured using ten-fold cross-validation. After fitting a model to 90\% of the data, we used the model to geolocate samples from the withheld 10\% and calculated prediction errors (the great-circle distance between a sample's true and predicted origin), prediction region coverage (the proportion of samples that are captured within a model's 90\% probability region), and prediction area matching (the proportion of predictions that lie in the same city or locality, county, if applicable, state or foreign equivalent, and country as that of the true sample point).

We analyzed the data at the national level, with samples distributed over the continental United States (U.S.), as well as a subset of these data at the regional level, with samples across the Research Triangle of central North Carolina (N.C.), encompassing the cities of Raleigh, Durham, and Chapel Hill. At the global level, we analyzed a new dataset of dust-associated fungal microbiome samples from countries located across Eastern Europe, Middle East, Africa, Asia, Oceania, and the Americas. 

\subsection{National}
	
We began with an analysis of the full samples from the continental U.S. (excluding samples from Alaska and Hawaii) with $n=1,301$ samples of $p=57,331$ distinct fungal phylotypes. Under the assumption that the sampling density should reflect population density across the U.S., we weight samples upward (downward) from states where the number of samples underrepresents (overrepresents) the state population relative to the national population. For instance, North Carolina contributes the most samples despite being the 9th most populous state, so the contribution of each individual N.C. point to the model is downweighted to better reflect density patterns nationwide. Specifically, training samples are weighted by the ratio of the state's population to the number of samples from the state.
%Specifically, let $A_t,\,t=1,\ldots,T$ represent the population areas (in this case continental U.S. states, $T=48$), $\text{pop}(A_t)$ is the population of $A_t$, and define $A(\bs) = A_t$ for $\bs\in A_t$. Then a sample point with origin $\bs_i$ is given weight $w_i = \left\{\text{pop}[A(\bs_i)]\,\middle/\,\sum_{t=1}^T\text{pop}(A_t)\right\} \cdot \left\{\sum_{j=1}^n I[\bs_j\in A(\bs_i)]\right\}^{-1}$.

\begin{table}\caption{Geolocation predictions within the continental United States using ambient dust-associated fungal microbiome data.  Methods are compared in terms of median absolute prediction error (``ME'') in kilometers, coverage (``COV'') of 90\% prediction regions, and classification accuracy for predicting the state, county and city.}
\begin{tabular}{llrrrrrr}\label{tab:national-cv}
& & \multicolumn{1}{c}{} & \multicolumn{1}{c}{} & \multicolumn{3}{c}{Area match (\%)} \\
			Seeds &      Model &   ME &     COV  &     State &  County &  City \\
			\hline
		   Coarse &  Spatial NN & 231.9 &   96.0 &   44.9 &    12.6 &  10.4 \\
			      &   Spatial RF & 194.8 &   99.2 &   48.3 &    13.9 &  11.9 \\
			      &   Spatial Net & 87.7  &   97.0 &   60.6 &    22.1 &  17.6 \\ \vspace{0.5em}
			      &    DeepSpace & 86.9  &   98.2 &   61.0 &    21.5 &  17.2 \\
			Fine  &  Spatial NN & 258.3 &   80.5 &   43.9 &    14.5 &  11.8 \\
			      &   Spatial RF & 221.5 &   96.5 &   46.8 &    17.2 &  14.8 \\
			      &   Spatial Net & 119.5 &  89.0 &   56.5 &    24.4 &  20.7 \\ \vspace{0.5em}
			      &    DeepSpace & 107.6 &  93.1 &   58.7 &    23.8 &  20.0 \\
			Mixed &  Spatial NN & 247.9 &  90.0 &   44.6 &    14.6 &  12.1 \\
			      &   Spatial RF & 213.7 &  98.6 &   47.6 &    17.0 &  14.2 \\
			      &   Spatial Net & 113.3 &  94.3 &   58.2 &    23.9 &  19.7 \\ \vspace{0.5em}
			      &    DeepSpace & 97.8  &  96.3 &   60.2 &    23.6 &  19.4 \\
			None  &          BDA & 263.7 &  91.0 &   31.9 &     1.6 &   0.8 \\
			      &  State DNN   & 211.0 &      - &   57.1 &     - &   - \\
		\end{tabular}
	\end{table}
	
Table~\ref{tab:national-cv} summarizes the performance of the competing geolocation models. For all three partitioning schemes, there was a reduction in prediction error as the complexity of the spatial classifier increases (NN to RF to Net to DS). The coarse partitioning scheme appeared to consistently lead to the lowest prediction error, but due to the large size of the cells, the coverage probabilities are over-inflated. The fine partitioning achieved slightly lower coverage probabilities at the expense of higher prediction errors. As might be expected, mixed partitioning strikes a balance between coarse and fine with respect to their differences in prediction error and probability region coverage. Among the spatial models, DS regularly outperformed NN, RF, and Net in predicting a sample's true state, county, and city of origin. In fact, DS appears better at predicting the state of origin for a sample than a DNN fit specifically for state classification. Thus, operating at the point-level appears to be beneficial, perhaps because it allows DS to learn regional patterns that help it distinguish between states.
	
\begin{figure}
\centering
\includegraphics[width=0.95\textwidth]{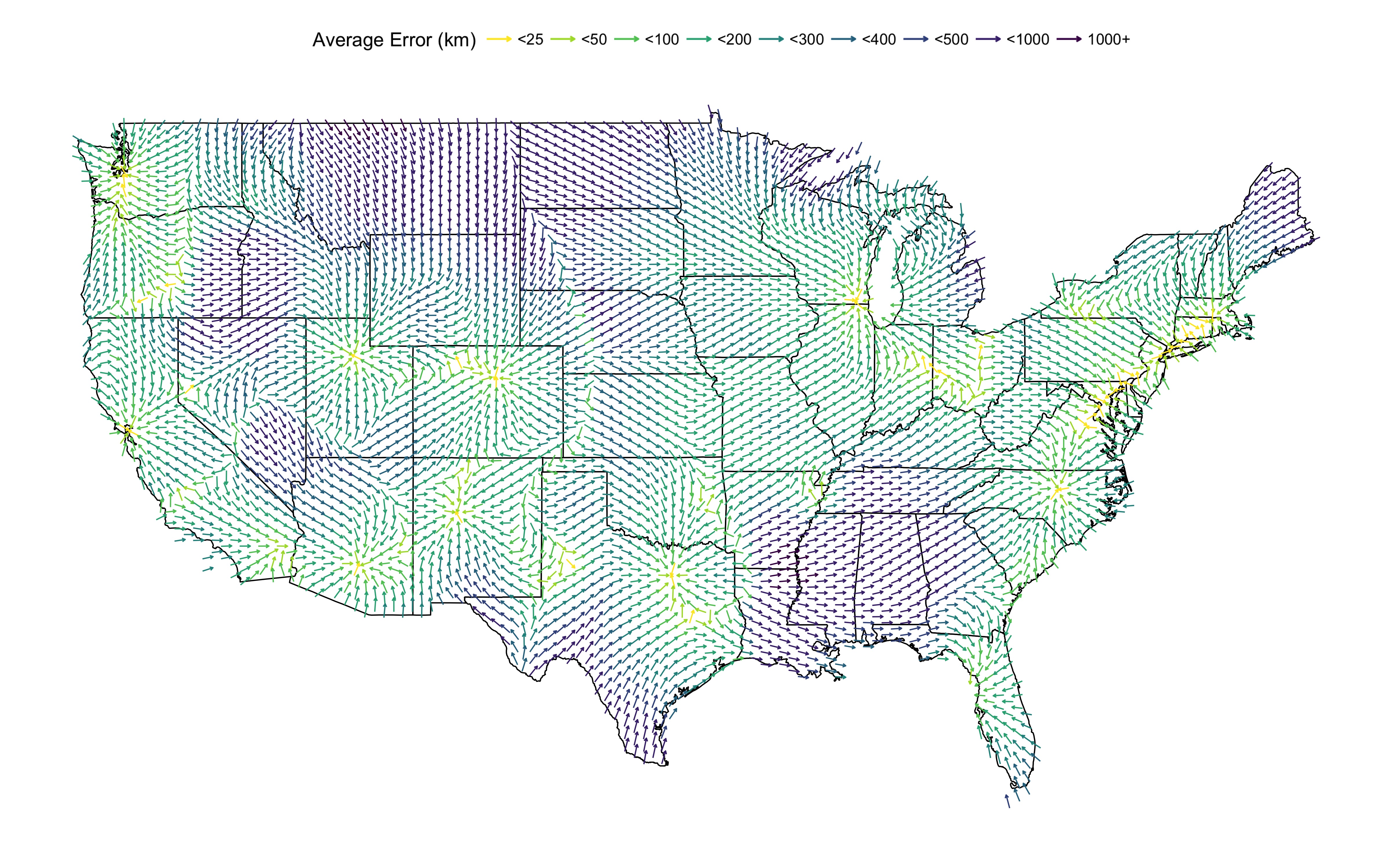}
\caption{Average prediction errors made by Coarse DS over ten-fold cross-validation. Each arrow points from a sample location in the direction to which a prediction is likely to be made, and the arrow's color indicates how far on average that prediction is likely to be from the true origin.}
		\label{fig:national-preds-ds}
	\end{figure}

Figure \ref{fig:national-preds-ds} depicts the average prediction errors made by DS under the coarse partitioning scheme.  In this plot, the arrows point to the average prediction direction and the color indicates the average prediction error.  The averages are computed using a Gaussian kernel smoother with a bandwidth of 100 km.  Predictions are biased towards populated urban areas for which more data are available than surrounding rural areas. DeepSpace has the lowest prediction error throughout the Northeast, Midwest, and Western United States. Despite downweighting points in North Carolina to better reflect population patterns nationwide, central N.C. attracts a large number of predictions away from samples collected in Virginia, Tennessee, and the Southern U.S. (with the exception of Florida).
	
%	\begin{figure}
%		\centering
%		\includegraphics[width=0.49\textwidth]{figures/national-preds-rf.png}
%		\captionof{figure}{Average prediction errors made by Coarse Spatial RF over ten-fold cross-validation. Each arrow points from a sample location in the direction to which a prediction is likely to be made, and the arrow's color indicates how far on average that prediction is likely to be from the true origin.}
%		\label{fig:national-preds-rf}
%	\end{figure}

\subsection{Regional}
	
We then analyzed a subset of the U.S. data, only $n=116$ samples with $p=20,557$ distinct fungal phylotypes belonging to the central North Carolina counties of Wake, Durham, and Orange.  At the national scale, differences in fungal occupancy are likely to reflect both biogeographical differences in terms of which taxa occur where and local habitat differences. When we perform analysis at this regional scale, the taxa pool of fungi that could colonize any particular site should be similar across all sites with any differences more likely driven by stochastic processes or differences in local environmental conditions. By focusing on a small geographic area we can isolate the ability of the models to predict the origin of a sample when biogeographic differences are held relatively constant. In this analysis we seek to determine if there is a limit to the resolution that one may geolocate samples using fungal occupancy data.

\begin{table}\caption{Geolocation predictions within the Triangle region of North Carolina using ambient dust-associated fungal microbiome data. Methods are compared in terms of median absolute prediction error (``ME'') in kilometers, coverage (``COV'') of 90\% prediction regions, and classification accuracy for predicting the county and city.}\label{tab:local-cv}
\centering
\begin{tabular}{llrrrr}
			& & \multicolumn{1}{c}{} & \multicolumn{1}{c}{} & \multicolumn{2}{c}{Area match (\%)} \\
			%\cmidrule(r){3-5} \cmidrule(r){6-8} \cmidrule(r){9-10}
			Seeds &      Model &   ME & COV &  County &  City \\
			\hline
			Coarse &  Spatial NN &  19.2 & 91.4 &   44.8 &  25.0 \\
			       &   Spatial RF &  17.4 & 98.3 &   38.8 &  29.3 \\
			       &   Spatial Net &  18.9 & 90.5 &   41.4 &  21.6 \\ \vspace{0.5em}
			       &    DeepSpace &  19.1 & 89.7 &   44.0 &  24.1 \\
			Fine   &  Spatial NN &  23.3 & 73.3 &   40.5 &  17.2 \\
			       &   Spatial RF &  19.5 & 87.9 &   36.2 &  16.4 \\
			       &   Spatial Net &  18.5 & 76.7 &   40.5 &  17.2 \\ \vspace{0.5em}
			       &    DeepSpace &  19.6 & 82.8 &   40.5 &  18.1 \\
			Mixed  &  Spatial NN &  20.4 & 84.5 &   43.1 &  24.1 \\
			       &   Spatial RF &  20.2 & 93.1 &   36.2 &  19.0 \\
			       &   Spatial Net &  19.2 & 90.5 &   49.1 &  25.9 \\ \vspace{0.5em}
			       &    DeepSpace &  20.0 & 90.5 &   40.5 &  18.1 \\
			None   &          BDA &  19.5 & 90.5 &   40.5 &  19.0 \\
			       &  County DNN  &  18.0 &    - &   53.4 &   - \\
		\end{tabular}
	\end{table}

\begin{figure}
\centering
\includegraphics[width=0.9\textwidth]{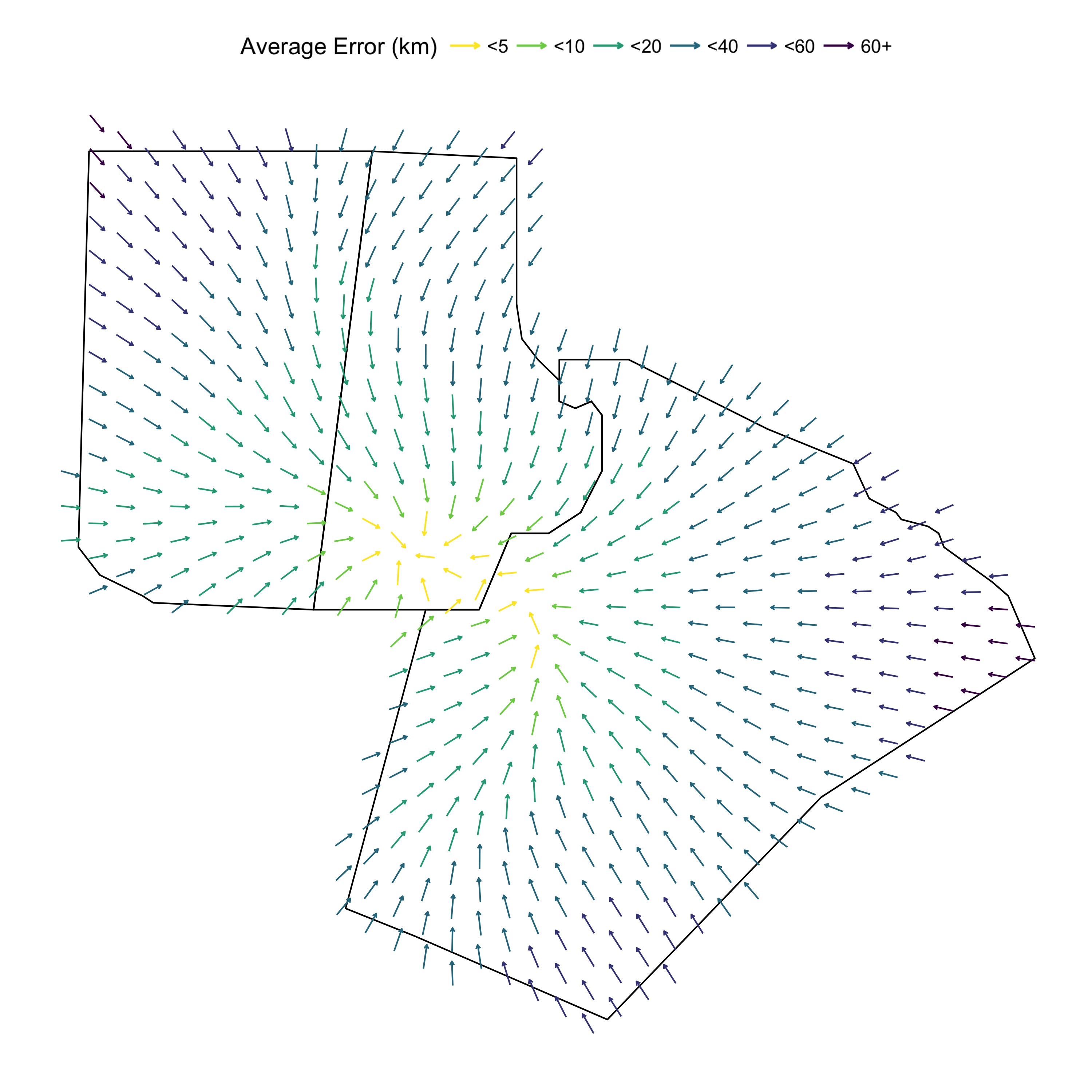}
\caption{Prediction errors made by Coarse DS over ten-fold cross-validation. Each arrow points in the direction to which a prediction is likely to be made, and the arrow's color indicates how far that prediction is likely to be from the true origin, on average.}
\label{fig:local-preds}
\end{figure}

Table~\ref{tab:local-cv} summarizes geolocation efforts over this localized region. As in the national level analysis, we weighted samples proportionally to population density, this time by county population. Unlike the national level analyses, many of the spatial models perform similarly across the three partitioning schemes. In fact, BDA and County DNN achieve the lowest prediction errors. Moreover, County DNN has the highest county classification accuracy (53.4\%) compared to the spatial models (36.2 to 49.1\%). The county classification rates for all models exceeds $1/3$, suggesting there is some hope of county-level prediction with these data. Figure~\ref{fig:local-preds} shows average prediction errors made by DeepSpace and County DNN (using a smoother bandwidth of 5 km), where the former makes most of its predictions between Durham and Raleigh, and the latter centers on Raleigh, the most populous city in the area. 
		
\subsection{Global}

Finally, we analyzed a global level dust-associated microbiome dataset with $n=399$ samples with $p=15,475$ distinct fungal phylotypes collected across 28 countries in Eastern Europe, Middle East, Africa, Asia, Oceania, and the Americas. There were $10-15$ sampling locations in each country.  %Due to this relative balance in per-country sample collection effort, we do not weight the samples to reflect country population. Moreover, 
Samples within each country often stem from a single major city and thus do not permit estimation of within-country variation.  We therefore compared the models only on their ability to detect a sample's country of origin. As was demonstrated in the national level analyses, DS is capable of learning regional patterns in the data which could improve its classification accuracy.

\begin{figure}
\centering
\includegraphics[width=0.75\textwidth]{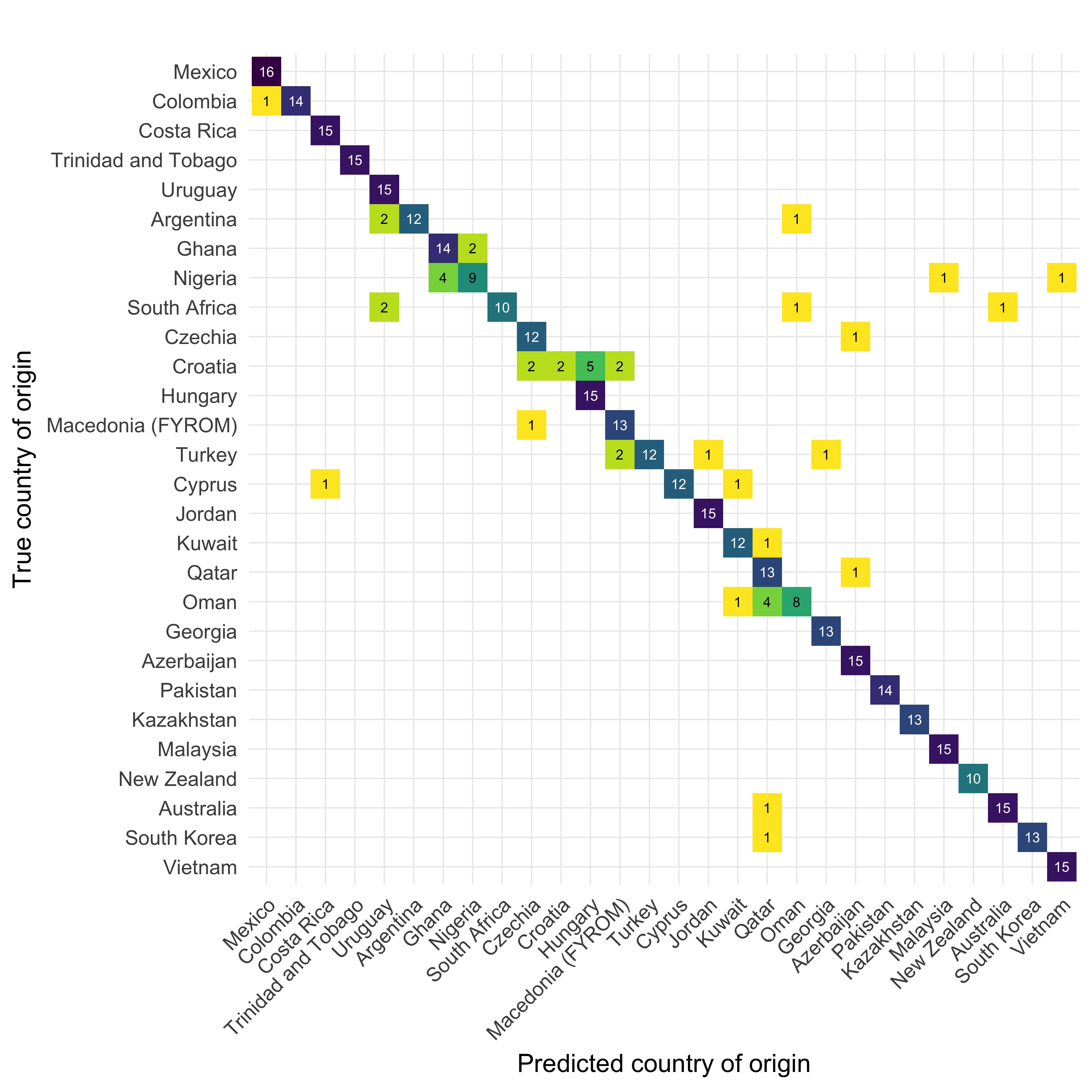}
\includegraphics[width=0.75\textwidth]{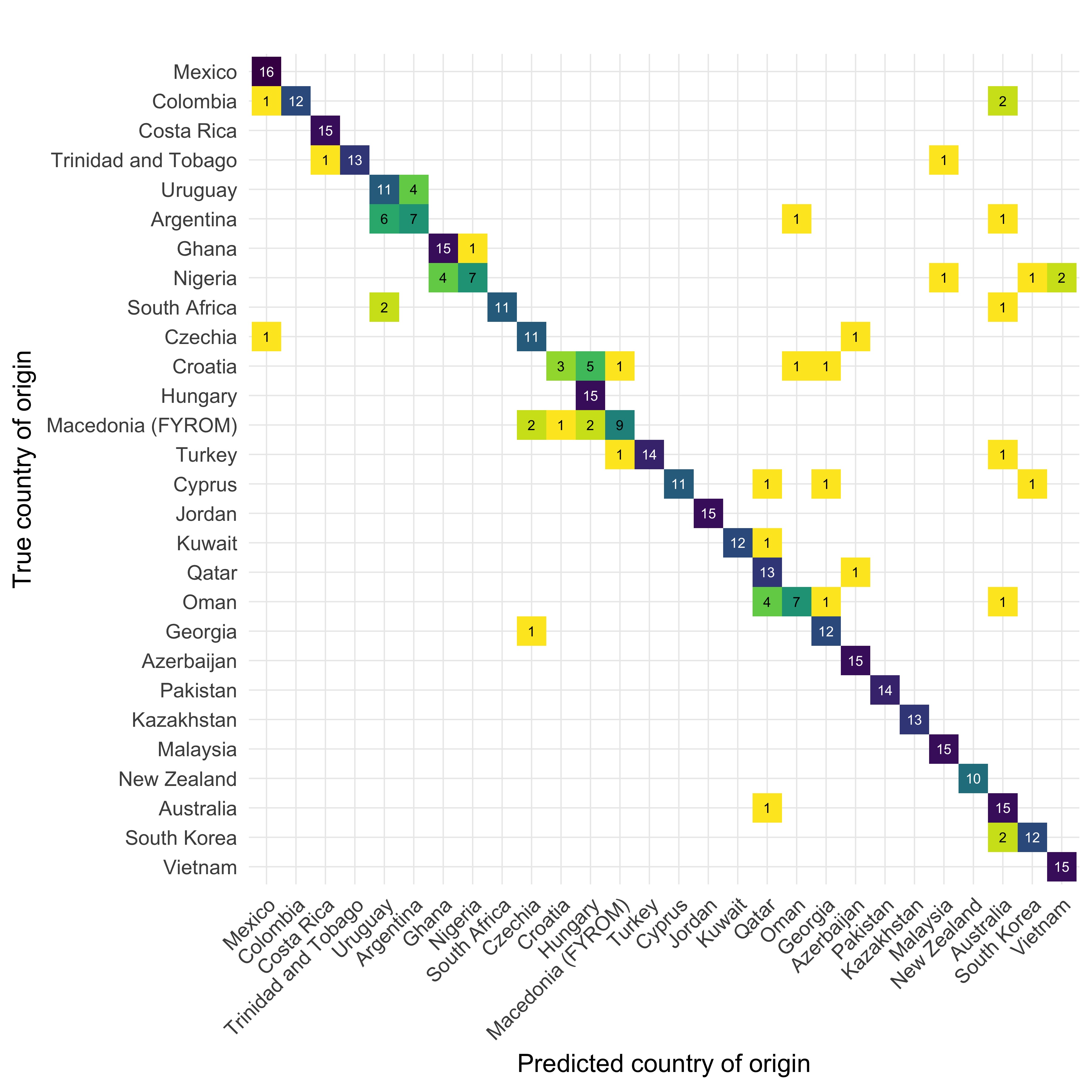}
\caption{Country classification between Mixed DS (top) and Country DNN (bottom). Each plot depicts the frequency of samples predicted to originate from a country ($x$-axis) against their true  origin ($y$-axis).}
\label{fig:country-confusion-matrix}
\end{figure}

Country DNN achieves a high classification rate of 84.7\%. Among the three partitioning schemes, spatial models perform best with the mixed partitions which achieve country classification rates of 62.7\% (Spatial KNN), 74.9\% (Spatial RF), 84.2\% (Spatial NN), and 89.5\% (DeepSpace). Figure~\ref{fig:country-confusion-matrix} depicts the true and predicted countries for the Mixed DS and Country DNN models. The models struggle to distinguish between samples from the bordering countries of Uruguay and Argentina, but DS only misclassifies 2 samples compared to 10 samples misclassified by DNN. 
	
\pagestyle{empty}

Samples for Croatia were often misclassified as being from nearby countries by DS. Interestingly, Macedonia, on the other hand, which is very near to Croatia, was almost perfectly classified by DS. DNN misclassified samples from both Croatia and Macedonia, but tended to classify them as being from nearby countries likely to share many fungal taxa with Croatia and Macedonia respectively. DS and DNN predicted 4 samples from Oman to be from Qatar, with DS mistakenly attributing an additional sample to Kuwait, and DNN attributing two samples to Georgia and, strangely, Australia. Overall, DS achieves noticeably fewer misses than DNN when classifying country, despite the latter model being trained specifically for this task. This suggests there are regional patterns within and between countries that a point-level model may exploit for higher accuracy forensic geolocations.

\section{Discussion}

We developed a new geolocation algorithm that combines random spatial partitions with deep learning classification. The DeepSpace algorithm exploits efficient deep learning software to nonparametrically estimate a prediction density as a function of an ultra-high dimensional predictor. DeepSpace was applied on three spatial scales: national, regional and global. The results of all geolocation methods were underwhelming in our regional analysis of the Triangle Region of North Carolina where, despite having over one hundred samples in these three counties, none of the methods achieves over a 53\% county classification accuracy. However, the DeepSpace results are outstanding on the national and global scales, with median prediction error less than 100 km in the continental U.S. and country classification nearly 90\% for the global analysis. In other words, given a sample of dust from anywhere in the U.S., we can identify the geographic origin of that sample to within 100 km on average. Given a sample of dust from anywhere in the world, we can identify the country of origin of that sample ninety percent of the time. 
	
A limitation of our analyses is that the sampling locations were not selected randomly or systematically. The U.S. data were contributed by volunteer citizen scientists and only uses samples from their door trims. For the global dataset the countries were carefully selected to represent the spatial domain of interest and the design was balanced across countries, but the sample locations within country were selected primarily based on convenience.  We accounted for these sampling issues in our analysis by weighting observations in the national analysis and limiting ourselves to country-level predictions in the global analysis, but the sampling scheme should still be considered when interpreting and generalizing our cross-validation results. Nonetheless, these data are more informative than any other data source we are aware of and the statistical contribution of the proposed geolocation tool should prove useful in other settings.
	
We used only fungal data in our analysis, and a promising area of future work is to combine fungal and bacterial data.  Exploratory analysis show the fungal data to be more informative than were bacterial data, but more sophisticated ways to incorporate bacterial data are worth exploring.  Another area of future work is to include features other than spatial coordinates in the predictions.  For example, if there are substantial differences in the microbiomes of residential and forested areas, then predictions could be refined based on the most likely land-use type.  Finally, we are working to include a temporal component in our analysis.  A first step is to include seasonality in our geolocation algorithm, although it is likely that fungi accumulate over many months and so seasonal detection would likely require a different sampling approach (i.e., not settled dust).  A more ambitious task is to geolocate an object that has moved in space and time.  For example, given a reliable map of the microbiome through a spatial region, it may be possible to infer that a package spent a significant amount of time in two locations by modeling its microbiome composition as a mixture of the microbiome composition at the two locations.

\subsection*{Acknowledgments}
Research was sponsored by the U.S. Army Research Office and the Defense Forensic Science Center (DFSC) and was accomplished under Cooperative Agreement Number W911NF-16-2-0195. The views and conclusions contained in this document are those of the authors and should not be interpreted as representing the official policies, either expressed or implied, of the Army Research Office, DFSC, or the U.S. Government. The U.S. Government is authorized to reproduce and distribute reprints for Government purposes notwithstanding any copyright notation hereon.  The work also was partially supported by National Science Foundation grants DMS-1513579 and DMS-1555141.

\bibliographystyle{rss}
\bibliography{DS}

\end{document}